\documentclass[conference]{IEEEtran}
\IEEEoverridecommandlockouts
\usepackage{cite}
\usepackage{multirow,subfigure,caption,amssymb,booktabs,xcolor}
\usepackage{arydshln}

\usepackage{amsmath,amssymb,amsfonts}
\usepackage{algorithmic}
\usepackage{graphicx}
\usepackage{textcomp}
\usepackage{xcolor}
\def\BibTeX{{\rm B\kern-.05em{\sc i\kern-.025em b}\kern-.08em
    T\kern-.1667em\lower.7ex\hbox{E}\kern-.125emX}}
\begin{document}

\makeatletter
\newcommand{\linebreakand}{%
  \end{@IEEEauthorhalign}
  \hfill\mbox{}\par
  \mbox{}\hfill\begin{@IEEEauthorhalign}
}
\makeatother

\newcommand{\modelname}{KAFT}

\title{
Knowledge Augmented Finetuning Matters 
in both RAG and Agent Based Dialog Systems
}

\author{
\IEEEauthorblockN{Yucheng Cai$^{*}$\thanks{ $^*$Equal contribution, $\dagger$Corresponding author.}}
\IEEEauthorblockA{\textit{EE Department, Tsinghua University} \\
Beijing, China \\
cyc22@mails.tsinghua.edu.cn
}
\and
\IEEEauthorblockN{Yuxuan Wu$^{*}$}
\IEEEauthorblockA{\textit{EE Department, Tsinghua University} \\
Beijing, China \\
wuyx23@mails.tsinghua.edu.cn}
\and
\IEEEauthorblockN{Yi Huang}
\IEEEauthorblockA{\textit{China Mobile Research Institute} \\
Beijing, China \\
huangyi@chinamobile.com}
\linebreakand
\IEEEauthorblockN{Junlan Feng}
\IEEEauthorblockA{\textit{China Mobile Research Institute} \\
Beijing, China \\
fengjunlan@chinamobile.com}
\and
\IEEEauthorblockN{Zhijian Ou$^{\dagger}$}
\IEEEauthorblockA{\textit{EE Department, Tsinghua University} \\
Beijing, China \\
ozj@tsinghua.edu.cn}

}

\maketitle

\begin{abstract}
Large language models (LLMs) have recently
been applied to dialog systems. 
Despite making progress, LLMs are prone to errors in knowledge-intensive scenarios. 
Recently, approaches based on retrieval augmented generation (RAG) and agent have emerged to improve the factual accuracy by enhancing the LLMs with knowledge retrieved from external knowledge bases (KBs).
This is mostly implemented by prompting the LLMs with instructions, examples and the retrieved knowledge.
However, LLMs may have difficulty using the retrieved knowledge effectively for response generation, because they are not well trained to do such generation for specific domains.
To mitigate this problem, we propose to finetune the LLMs in the RAG-based and agent-based systems with domain-specific data, together with domain-specific external knowledge, which is called \textbf{k}nowledge \textbf{a}ugmented \textbf{f}ine\textbf{t}uning (KAFT).
We base our study on the MobileCS2 dataset, a real-life customer service dialog dataset that features intensive knowledge interactions, to systematically compare the prompting and KAFT techniques in the RAG-based and agent-based systems. 
Experiment results show that \modelname{} substantially surpasses prompting in both RAG and agent systems, particularly in terms of factual accuracy.
To the best of our knowledge, this paper represents the first solid empirical work to investigate the KAFT idea.
\end{abstract}

\section{Introduction}
Recent progress in large language models (LLMs), such as GPT4 and PALM \cite{achiam2023gpt,chowdhery2023palm}, has shown improved performance in a range of natural language processing (NLP) tasks. 
These improvements have stimulated researchers and practitioners to integrate LLMs into real-world applications such as dialog systems.
For 
real-life dialog systems, it is crucial for LLMs to respond accurately and reliably, 
which usually require domain-specific knowledge. 
Despite their power, in knowledge-intensive dialog systems, LLMs often generate outputs that are inaccurate or misleading, a phenomenon known as ``hallucination'' \cite{ji2023survey}. This poses a significant challenge to the factual accuracy of the systems.

In order to mitigate the phenomenon of hallucination, several approaches have been proposed. %
Among them, the 
RAG approach stands out as a promising solution. 
By integrating knowledge retrieval into the generative system, RAG significantly enhances factual accuracy and reduces hallucination \cite{lewis2020retrieval, izacard2022atlas, cai2023knowledge}. 
In addition, there are growing interests in the agent-based approach, which exploits the tool-calling capability of LLM \cite{park2023generative,schick2024toolformer}. By employing API calls, the agent approach aims to improve factual accuracy within question-answering (QA and dialogue systems, as demonstrated by the study in \cite{yaoreact}. The knowledge obtained by the API calls in the agent-based system is similar to the retrieved knowledge in the RAG-based system. 

For both the RAG and agent based dialog systems, 
the commonly adopted implementation is to prompt the LLM to directly utilize the external knowledge obtained by the system as in Figure \ref{fig:main}(b). Instructions and examples are added to the prompts, along with the retrieved knowledge, to improve the system performance \cite{nori2023generalist}.
However, even with the instructions and examples, the LLMs may still have difficulty in effectively using the retrieved knowledge, because they are not well trained to do such generation for specific domains\cite{chen2023fireact,gao2024efficient}. For example, in the case shown in Figure \ref{fig:main}(b), the LLM does not understand the term ``directional flow'' which is specific to the mobile service domain. Therefore, the LLM cannot deduce that the reason for the user's overage flow is the use of the flow in other apps. 
To mitigate this problem, we propose to finetune the LLMs in the RAG and agent based systems with domain-specific data, together with the domain-specific external knowledge, which is called knowledge augmented finetuning (\modelname{}) in this work, as shown in Figure \ref{fig:main}(c). 
Conventionally, to adapt LLMs to a specific domain, the LLMs can also been directly finetuned, without using the RAG or agent-based systems, as shown in Figure \ref{fig:main}(a).
However, the performance of this direct finetuning of LLMs is often even inferior to the un-finetuned RAG-based systems \cite{gupta2024rag,soudani2024fine}. 
\emph{In this paper, we focus on the RAG and agent based systems, investigate the method of finetuning the LLMs with retrieved knowledge in those systems, and compare it to the method of prompting the LLMs in those systems.}

The idea of finetuning LLMs is not new, but prior works mostly study direct inference tasks with LLMs (such as close-book QA, text completion, machine translation, and so on). As far as we know, there is no solid work to investigate the KAFT idea in the knowledge-intensive tasks where RAG or agent based methods are built to retrieve knowledges from external KBs.
This paper represents the first solid empirical work to investigate the KAFT idea.
Futuremore, unlike other fine-tuning techniques such as the Lora technique \cite{hu2022lora}, which aims to improve the effectiveness and efficiency of fine-tuning, the proposed KAFT method aims to teach LLMs a new skill, i.e., the ability to make use of domain-specific external knowledge. 
KAFT teaches LLMs to take advantage of retrieved knowledge by constructing corresponding training data to finetune the model, similar to the instruction-tuning technique \cite{chung2024scaling} which teaches LLMs the ability to follow instructions.

\begin{figure}
    \centering
    \includegraphics[width=1.0\linewidth]{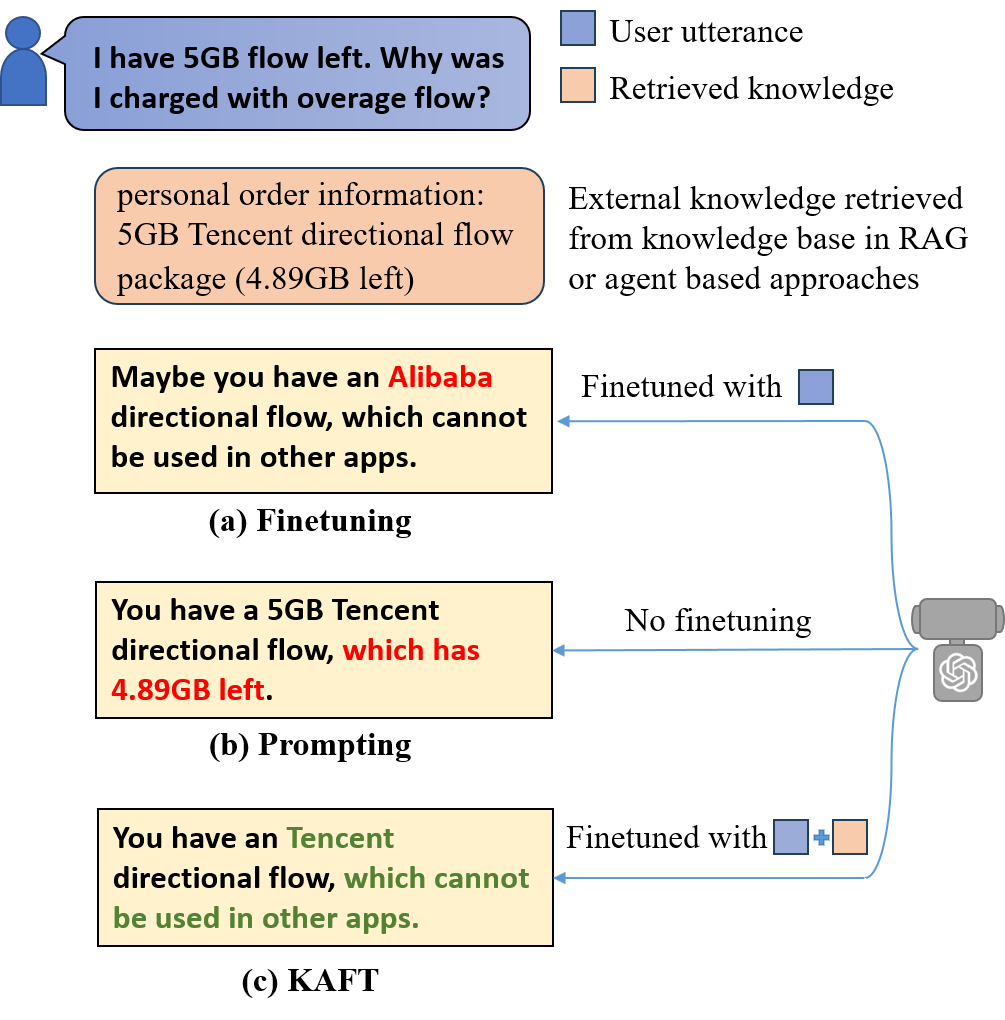}
    \caption{Overview of the three methods to improve factual accuracy for the LLM based dialog systems. Direct finetuning without knowledge still often leads to serious hallucination, as the LLM may be unaware of the information gap during the training and testing situations. Meanwhile, for the prompting method, the LLM is not trained on domain specific data, which cannot fully leverage the domain-specific external knowledge (such as the ``directional flow'' in the example). The proposed KAFT method finetunes the LLM on domain specific data with external knowledge, which overcomes the drawback of the direct finetuning method and the prompting method. }
    \label{fig:main}
\end{figure}

To demonstrate the efficacy of the proposed \modelname{} method, we build both RAG-based and agent-based dialog systems upon the MobileCS2 dataset \cite{cai20242nd}. 
The mobileCS2 is a real-life human to human knowledge-grounded dialog dataset, 
released from the SLT 2024 FutureDial Challenge \cite{cai20242nd}. 
It consists of real-world dialog transcripts between real users and customer service staff, along with annotated knowledge pieces necessary for staff to respond properly. Extensive experiments are conducted on the dataset to compare the \modelname{} method and the prompting method. 
The experiment results demonstrate that the 
\modelname{} technique 
can substantially outperform the prompting technique in both the RAG and the agent based systems, 
thereby showing the efficacy of the proposed \modelname{} method in knowledge-intensive dialog systems. 

In summary, the main contributions of this work are: 
\begin{itemize}
\item This paper proposes to teach the LLMs to make use of external knowledge by  finetuning the LLMs 
with domain-specific data, together with the domain-specific external knowledge, which is called knowledge augmented finetuning (\modelname{}). 

\item To validate the efficacy of the proposed \modelname{} method, RAG-based and agent-based dialog systems are built on the real-life customer service dataset MobileCS2 dataset to compare the proposed \modelname{} method with the prompting method. 

\item Extensive experiments on the MobileCS2 dataset show that the proposed \modelname{} method can improve the ability of LLMs to make use of knowledge and 
substantially surpass the prompting method in both RAG-based and agent-based dialog systems.
\end{itemize}

\section{Related Work}
\label{related}

\subsection{Large Language Models (LLMs)}

LLMs are large foundation models pretrained with corpus of trillions of tokens. 
The emergence of LLMs \cite{achiam2023gpt,chowdhery2023palm}
has greatly improved the performance in various NLP tasks. 
Previous studies have discovered the strong 
in-context learning \cite{brown2020language} and reasoning \cite{wei2022chain} ability of LLMs, which inspired the researchers to explore LLMs in more complicated tasks like question-answering and dialog systems \cite{ouyang2022training,touvron2023llama}. Despite their success in open-domain dialogs, the absence of specific domain knowledge and up-to-date facts in the data can pose limitations for those systems in vertical domains. 
The RAG-based approach \cite{lewis2020retrieval,izacard2022atlas,cai2023knowledge} and the agent-based approach \cite{park2023generative,schick2024toolformer} are used to mitigate this issue. 
The most commonly adopted implementation of the RAG-based system and agent-based system is to prompt the LLM with instructions and examples, along with the retrieved knowledge \cite{yaoreact}. However the LLMs still struggles to effectively utilize the knowledge in the RAG-based and agent-based systems as they lack the background information related to the specific areas \cite{chen2023fireact,gao2024efficient}. 
In this work, we propose to use the \modelname{} method to finetune the LLMs to gain the ability to make full use of the external knowledge. 

\subsection{Retrieval Augmented Generation}

Retrieval Augmented Generation (RAG) \cite{lewis2020retrieval} is a technique that enhances the performance of LLMs by utilizing external pre-stored data, such as texts, dialogues, and knowledge bases. Specifically, when the model needs to generate text or provide an answer, it can first retrieve relevant information from external sources and then generate more accurate and enriched outputs by integrating the information retrieved.
Therefore, a retrieval augmented generation system typically contains two components, a retriever and an LLM (also called a generator). There are several works that improve the original RAG work, mainly focused on improving the retriever \cite{karpukhin-etal-2020-dense,glass2022re2g,izacard2022contriever} and the generator \cite{guu2020realm,zhang2024raft,asai2024self,khattab2022demonstrate}. 
Unlike the previous works that focus on general domain question answering, this paper represents the first solid empirical work to investigate the idea that the ability of LLMs to make use of domain-specific knowledge can be improved by \modelname{}.

Recently, there are some studies to compare RAG with the method of directly finetuning LLMs without using RAG for vertical domains \cite{soudani2024fine,gupta2024rag}. Unlike those studies, this paper aims to compare the proposed \modelname{} method with 
the method of prompting the LLMs in both RAG-based and agent-based dialog systems to show that the ability of LLMs to make use of domain-specific knowledge can be enhanced by post-training.

\subsection{Large Language Model based Agents}

With the development of LLMs, recent researches have explored the potential of building agents upon LLMs, leveraging their strong generation and understanding abilities. 
The introduction of LLM based agents has significantly enhanced the ability of machines to interact with the world \cite{park2023generative}. Using the superior ability of the LLMs, those agents can plan their actions and interact with tools as human \cite{schick2024toolformer,touvron2023llama}. For knowledge-intensive tasks, the ability to interact with tools is important, as the agent can actively get the information necessary for accomplishing the task like human does. In this work, we explore the possibility of building a customer service agent that can act like real-life customer service staffs. 

\section{Method}

\subsection{Task and Definition}

In a customer service dialog system, 
assume we have a dialog $X$ 
with $T$ turns of user utterances and system responses, denoted by $u_{1},r_{1},\cdots,u_{T},r_{T}$ respectively. At turn $t$, based on the dialog context $c_{t} \triangleq u_{1} \oplus r_{1} \oplus \cdots \oplus u_{t-1} \oplus r_{t-1} 
\oplus  u_{t}$ ($\oplus$ means sequence concatenation), the system needs to 
generate an appropriate response leveraging the knowledge base (KB).
For the MobileCS2 dataset, 
the knowledge base is made up of the user information ($KB_{user}$), which is unique for each dialog, the product information list ($KB_{product}$), and the FAQ list for commonly asked questions ($KB_{FAQ}$). 

\subsection{Retrieval Augmentation Generation (RAG) based Dialog System}

\label{sec:rag}

\subsubsection{Knowledge Augmented Finetuning (\modelname{}) of LLMs}

\begin{figure}[t]
\centering
	\includegraphics[width=0.9\linewidth]{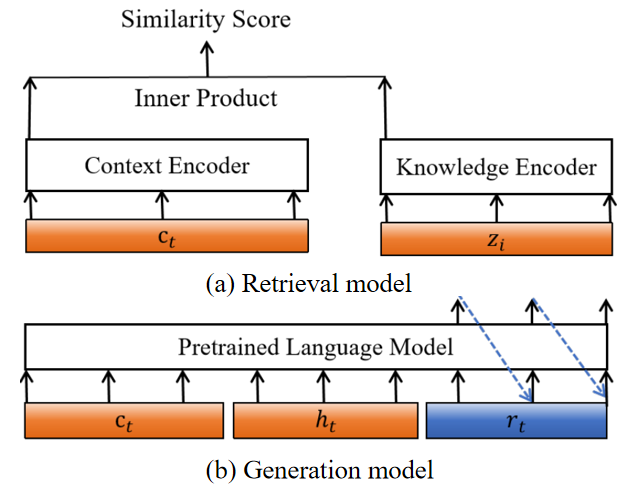}
 	
	\caption{Overview of the RAG-based dialog systems:  (a) the retrieval model, (b) the generation model.}
	\label{fig:rag}
\end{figure}

The RAG based dialog system employs a retriever to retrieve the relevant knowledge pieces from the knowledge base and uses the retrieved knowledge to help the generator (i.e. the LLM) generate the response. In this work, the RAG-based dialog system is similar to the system in \cite{cai20242nd}. For a dialog $X$ in the MobileCS2 dataset, the knowledge base $KB_{X}$ can be denoted as: 
$KB_{X} \triangleq KB_{user}\cup KB_{FAQ}\cup KB_{product}$. 
Given the knowledge base $KB_{X}$, at turn $t$ of a dialog $X$, the system uses a retriever $p_\eta(z_{i} \mid c_{t})$, which is shown in Figure \ref{fig:rag}(a), 
to obtain the relevant knowledge $h_{t}$ from the knowledge base and generate appropriate responses with the generator $p_\theta(r_{t} \mid c_{t},h_{t})$, which is shown in Figure \ref{fig:rag}(b).

The retriever is implemented with the dual-encoder architecture including the knowledge \underline{p}iece encoder $\operatorname{Encoder}_p$ and the \underline{c}ontext encoder $\operatorname{Encoder}_c $, as shown in Figure \ref{fig:rag}(a). 
To train the retrieval model, 
for each knowledge piece $z_i~(i=1, 2, \cdots, K)$ in $KB_{X}$, the models fit the retrieval distribution of $p_\eta(z_{i}  \mid c_{t})$ as in   \cite{lewis2020retrieval}:  
\begin{align}
p_\eta(z_{i} \mid c_{t}) \propto \exp \left(\operatorname{Encoder}_p(z_{i})^{\top} \operatorname{Encoder}_c (c_{t})\right) 
\end{align}
$\operatorname{Encoder}_p$ and $\operatorname{Encoder}_c $ are both initialized with a BERT-based pretrained model \cite{devlin2019bert}. 
The log probabilities of the positive pieces $z \in Z_{+}$ (labeled in the dataset) are optimized: 
\begin{equation}
\mathcal{L}_{ret}
=-\frac{1}{\mid Z_{+} \mid}\sum_{z \in Z_{+}} \log p_\eta (z \mid c_{t})
\label{eq:retriever}
\end{equation}
The $\operatorname{Encoder}_p$ is fixed during the training, while the 
$\operatorname{Encoder}_c$ is trained with the loss in Eq. \ref{eq:retriever}, following the setting in \cite{karpukhin-etal-2020-dense}.

\textbf{KAFT for RAG} refers to finetune the LLM with the dialog context and knowledge piece, i.e., to finetune the generation model $p_\theta(r_{t} \mid c_{t},h_{t})$.
We use the auto-regressive loss to optimize the generation probability: 
\begin{align}
{p}_{\theta}(r_{t} \mid c_{t},h_{t})  
=\prod_{l=1}^{|r_{t}|}
p_{\theta}(y^l \mid c_{t}, h_{t}, y^1, \ldots, y^{l-1}) 
\label{eq:gpt2}
\end{align}
where $|\cdot|$ denotes the length in tokens, and $y^l$ the $l$-th token of $r_t$. This is similar to \cite{cai20242nd}. 
The baseline system in \cite{cai20242nd} used oracle knowledge in training the LLM.
However, we use the generated $h_{t}$ from the retriever rather than the annotated $h_{t}$ so that the training procedure of the generation model is aligned to the test setting where the annotated $h_{t}$ is not available. This adaption, similar to the noise adding technique in \cite{zhang2024raft}, brings some improvement to the system in our experiments. The generation model in Eq. \ref{eq:gpt2} is initialized with a GPT2-based pretrained language model \cite{radford2019gpt2}. 

\subsubsection{Prompting of LLMs}

In the method of \textbf{prompting LLMs for RAG}, we use the prompted LLM as the generator, while using the same retriever as in KAFT.
The in-context learning (ICL) \cite{brown2020language} method is used to prompt the LLM to generate appropriate responses given the examples randomly selected from the dataset. 
The generation probability of the LLM can be written as ${p}_{\theta}(r_{t} \mid prompt_{t},c_{t},h_{t})$. The prompt $prompt_{t}$ contains the instruction for the LLM and the example dialogs, which clearly instructs the LLM to generate the response leveraging the retrieved knowledge $h_t$. The prompt for the LLM, as well as an example of a turn in a dialog using the RAG-based system is shown in Figure \ref{fig:llm_prompt}(a).


\subsection{Agent based Dialog System}

\begin{figure}[t]
\centering
	\includegraphics[width=1\linewidth]{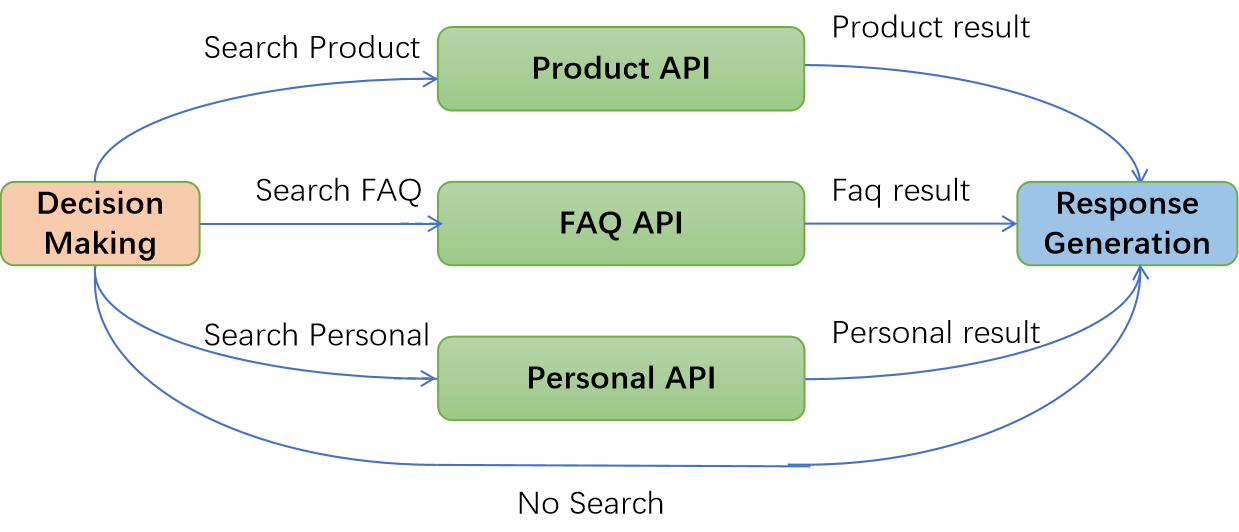}
	\caption{Overview of the agent-based dialog systems. The system first decides the search intention, then calls the corresponding API to perform the search operation. The system then  generates the response based on the search results.}
	\label{fig:agent}
\end{figure}

\begin{figure*}[t]
    \centering
    \includegraphics[width=1.0\linewidth]{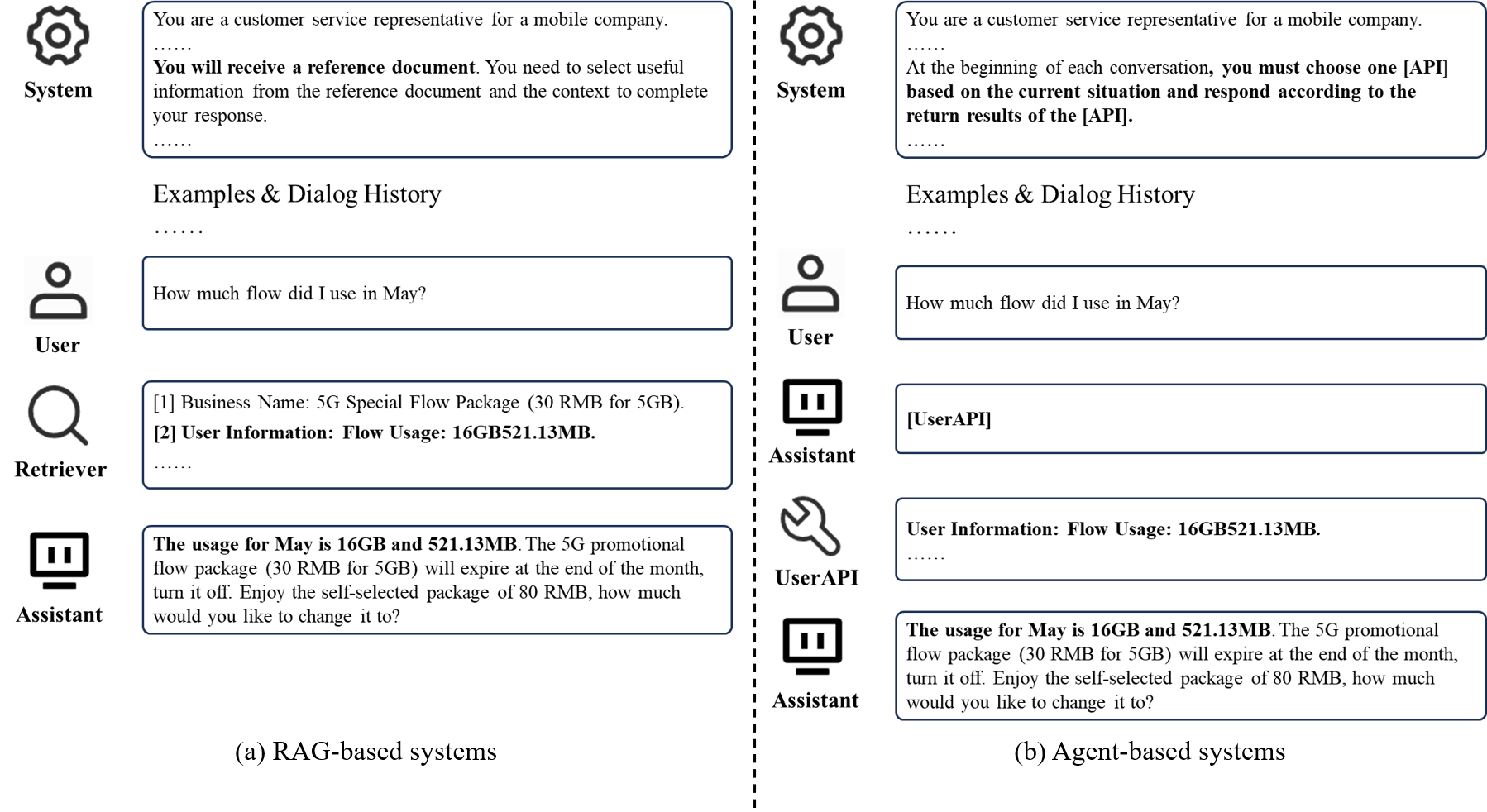}
    \caption{ An illustration of the prompts for the LLMs, as well as an example turn in the dialog in the RAG-based and agent-based systems. }
    \vspace{-0.5em}
    \label{fig:llm_prompt}
\end{figure*}

\subsubsection{Knowledge Augmented Finetuning (\modelname{}) of LLMs}
The agent based dialog system leverages the planning and search ability of the agent to accomplish the dialog, as shown in Figure \ref{fig:agent}. The agent consists of a decision maker, API calling, and an LLM as the generator.
At turn $t$ in a dialog, the agent first makes the search decision $a_{t}$ of what database the system needs to search based on the input context $c_t$. 
Based on the decision $a_{t}$, the agent conducts the corresponding search. 
If $a_{t}$ is `No Search', then the agent responds directly to the context; otherwise, the agent queries the corresponding API for the product, FAQ, and personal information.

To simulate the Product API, the FAQ API, and the Personal APIs, we use the annotated data of the corresponding search decisions in the MobileCS2 dataset. For example, the turns annotated with the search decision of ``Search Product'' are used to train the Product API. Notably, the purposes of all these search APIs are to retrieve some knowledge pieces from the knowledge bases, which is similar to the retrieval in RAG. Thus, for building the Product API and the FAQ API, we separately train two dual-encoder based retrievers, using the architecture in Figure \ref{fig:rag}(a) and the loss function in Eq. \ref{eq:retriever}.
For the personal API, we return all the personal information to ensure the recall of the information, as the knowledge base for the user information ($KB_{user}$) is relatively smaller than other knowledge bases.

The search result can be viewed as $h_{t}$, denoting a kind of knowledge piece.

\textbf{KAFT for agent} consists of finetuning LLMs for response generation given $h_t$ and decision making given $a_t$, respectively.
We use the similar auto-regressive loss as in KAFT for RAG to optimize the generation probability $p_\theta(r_{t} \mid c_{t},h_{t})$ (Eq. \ref{eq:gpt2}). 
The difference is that $h_{t}$ in the agent is the result given by the API search rather than from the retriever in RAG.
The decision maker of the agent is finetuned based on the probability ${p}_{\theta}(a_{t} \mid c_{t})$, also using auto-regressive loss, where the decision $a_{t}$, as a token sequence, can be viewed as another kind of knowledge.

\subsubsection{Prompting of LLMs}

In the method of \textbf{prompting LLMs for agent}, we employ the prompted LLM to implement decision making and response generation, while using the same search API as in KAFT.
We use the ICL) method \cite{brown2020language} to prompt the LLM to generate appropriate search decisions and responses, given the contexts and the prompts. We give corresponding examples and instructions, which clearly instruct the LLM which task, decision making or response generation, to perform. 
As there are only four possible search decisions, we enumerate them in the prompts for the decision making and ask the LLM to choose one of the 4 search decisions in the decision making task. The prompt for the LLM, as well as an example of a turn in a dialog using the agent-based system is shown in Figure \ref{fig:llm_prompt}(b).

\vspace{-0.5em}
\section{Experiment}
	
\subsection{Experiment Settings }

\begin{table*}[t] \large
\caption{Comparison between the dialog systems built with different methods and settings 
on the MobileCS2 dataset. 
``Direct respond'' means that we do not use the knowledge base and let the system to directly respond given the context.}

\begin{center}
\begin{tabular}{cccccc}
\toprule
Method 
& Setting &BLEU & BERTScore &Inform & Combined Score\\
\midrule
\multirow{3}{*}{Direct Respond }
& prompt + 0-shot (GPT3.5) & 4.81 & 0.601 & 0.003 & 0.328 \\
& prompt + 5-shot (GPT3.5)   &  11.4
& 0.646 & 0.002 & 0.382 \\
\cdashline{2-6}
&   
finetuning (GPT2) &  \textbf{17.3} 
& \textbf{0.652}&  \textbf{0.011} & \textbf{0.424}  \\
\midrule
\multirow{4}{*}{RAG}
& prompt + 0-shot (GPT3.5) & 17.2
& 0.657 & 0.063 & 0.478 \\
 &  prompt + 5-shot (GPT3.5)   &  20.6 
 & 0.663 & 0.059 & 0.493 \\
\cdashline{2-6}
 &  \modelname{} (GPT2) 
 & \textbf{22.2}
  & \textbf{0.668} & \textbf{0.145} & \textbf{0.590} \\
\midrule
\multirow{3}{*}{Agent}
& prompt + 0-shot (GPT3.5)
& 10.4 
& 0.620 & 0.033 & 0.395\\
 &  prompt + 5-shot (GPT3.5)
 &  18.0 
 & 0.645 & 0.082 & 0.495 \\
\cdashline{2-6} 
&  \modelname{} (GPT2) 
& \textbf{23.6} 
  & \textbf{0.656} & \textbf{0.147} & \textbf{0.594} \\
    \bottomrule
    \end{tabular}
    \end{center}
    \label{tab:main}
    \vspace{-0.5em}
\end{table*}

The experiments are carried out on a real-life human-human dialog dataset, called MobileCS2, released from the SLT 2024 FutureDial-RAG Challenge \cite{cai20242nd}. 
The MobileCS2 dataset is derived from the real-world mobile conversational scenarios and comprises around 3000 carefully annotated dialog logs between customers and customer service staffs. 
The dataset aims to promote the study of training and testing dialog systems for knowledge-intensive customer service. 
The dataset was officially  split into train, development and test sets, which consist of 1,926, 412 and 413 dialog samples, respectively. 

For evaluation, we follow the official scripts in \cite{cai20242nd}. 
To evaluate the retriever in RAG and the search APIs in agent, we use the recall metrics and report recall@1, recall@5 and recall@20. To evaluate the whole dialog system, we use three metrics. The generated response is evaluated by measuring the similarity score with the ground truth response (\emph{BLEU} and \emph{BERTScore}) and whether the system correctly provides the requested information by the user (\emph{Inform Rate}).
\emph{BLEU} measures the fluency of the generated responses by analyzing the amount of n-gram overlap between the real responses and the generated responses. 
\emph{BERTScore} \cite{zhang2019bertscore} measures the semantic similarity of the generated responses with the oracle responses by using a pretrained BERT model. 
\emph{Inform Rate} refers to how often the system response is able to cover the information requested by the user. 
The final score is calculated as $score = 0.5*(BLEU/100 + BERTScore) + Inform$, as in the original scripts in \cite{cai20242nd}. 

For the \modelname{} method, 
we finetune the GPT2 \cite{radford2019gpt2} in this study, while for the prompting 
method, we use the GPT3.5 \cite{ouyang2022training}. 
In the experiments, hyperparameters are chosen based on the development set and evaluated on the test set.

\subsection{Main Results}
\label{sec:main}



In the experiments, we examine the efficacy of the \modelname{} method compared to the prompting method. 
Based on the results in Table \ref{tab:main}, we find that \modelname{} 
can greatly boost the performance over prompting in both RAG-based and agent-based systems. While prompting the LLM with 5 examples can improve the performance, the dialog systems built with the prompting method 
still lag behind the systems built with the \modelname{} method on the BLEU, BERTScore, Inform and the Score metrics, especially on the Inform metric that requires accurate understanding and utilization of the domain-specific knowledge. The results demonstrate that the proposed \modelname{} method can substantially improve the ability of LLMs to make use of knowledge.

Moreover, according to the results in Table \ref{tab:api_predict} for agent-based systems, prompting the LLM with examples and instructions cannot perform well in the decision-making task, which shows a large performance gap behind the \modelname{} method. Presumably, this is because the complex contexts in real-life customer service dialogs make it difficult for the LLM to accurately predict the search decision given only 
instructions and examples. 

Overall, these results show that by using the \modelname{} method, the system can be greatly improved on both the response quality and the factual accuracy, mainly because the system is trained to adapt to speaking tunes and thinking manner for the vertical domain. 
This finding reflects the importance of the proposed \modelname{} method 
for building dialog systems for vertical domains.


Note that in the experiments, the KAFT method is implemented with GPT2, which is small.
Also note that the main research question investigated in this paper is to systematically compare the prompting and KAFT techniques for the RAG-based and agent-based systems.
It is found in our experiments that a small model like GPT2 with KAFT can beat GPT3.5 with prompting in the knowledge-intensive vertical domain, which clearly shows the advantage of the KAFT method. 
Using GPT-2 suffices to investigate the research question.
\begin{table}[t] \tiny
\caption{The decision making accuracy in the agent-based system
for the Personal, Product and FAQ search, using the prompting method and the \modelname{} method respectively.}
		\centering
		\resizebox{\linewidth}{!}{
			\begin{tabular}{cccc}
				\toprule
				
	Setting &Personal &Product &FAQ   \\
				\midrule
   0-shot (+prompt)  &0.183 &0.357  &0.005 \\
5-shot (+prompt)  &0.290  &  0.468 & 0.355  \\
\cdashline{1-4}
\modelname{} &\textbf{0.381}  &  \textbf{0.580}&\textbf{0.475}  \\
         \bottomrule	
	\end{tabular}}

	\label{tab:api_predict}
\end{table}



\begin{table}[t]
\caption{Comparison of the retrieval performance between the search APIs in the agent system and the retriever in the RAG system, for the Product search and FAQ search tasks. }
		\centering
		\resizebox{\linewidth}{!}{
			\begin{tabular}{ccccc}
				\toprule
				
	Task &Model &Recall@1 &Recall@5 &Recall@20  \\
				\midrule
   \multirow{2}{*}{Product Search} & Retriever &0.049 &0.132  &0.398 \\
&Product API &\textbf{0.075}  &  \textbf{0.199}&\textbf{0.451}  \\
        \midrule
      \multirow{2}{*}{FAQ Search} & Retriever &0.395  &0.649  &0.782 \\
&FAQ API &\textbf{0.546}  &  \textbf{0.782}&\textbf{0.872}  \\   
         \bottomrule	
	\end{tabular}}
 
	\label{tab:ret_result}
\end{table}

\begin{table}[t] 
\tiny

\caption{Ablation study about using the retrieved knowledge pieces (denoted by ``Retrieve'') versus using the annotated knowledge pieces (denoted by ``Oracle'') in \textbf{testing} in the RAG-based system (using retrieved knowlede in training).}
		\centering
		\resizebox{\linewidth}{!}{
			\begin{tabular}{ccccc}
				\toprule
				
	Test setting &BLEU &BERTScore &Inform &Score  \\
				\midrule
   Retrieve & 22.23 &0.668 &0.145  &0.590 \\
Oracle &\textbf{48.03} &\textbf{0.720}  &  \textbf{0.392}&\textbf{0.992}  \\ 
         \bottomrule	
	\end{tabular}}
 
	\label{tab:oracle_result}
\end{table}

\begin{table}[t] \tiny

\caption{Ablation study about using the retrieved knowledge pieces (denoted by ``Retrieve'') versus using the annotated knowledge pieces (denoted by ``Oracle'') in \textbf{training} in the RAG-based system  (using retrieved knowlede in testing).}
		\centering
		\resizebox{\linewidth}{!}{
			\begin{tabular}{ccccc}
				\toprule
				
	Test setting &BLEU &BERTScore &Inform &Score  \\
				\midrule
   Oracle &14.09 &0.640  & 0.127 & 0.517 \\
Retrieve &\textbf{22.23} &\textbf{0.668}  &  \textbf{0.145}&\textbf{0.590}  \\ 
         \bottomrule	
	\end{tabular}}
 
	\label{tab:noise_result}
\end{table}

\subsection{Analysis and Ablation}

\label{sec:ablation}
				
 
As shown in Table \ref{tab:main}, both the RAG based and the agent based systems show great improvements over the systems that directly respond to the user given the context, in terms of all the BLEU, BERTScore, Inform and the Score metrics.
This finding 
shows that both the RAG and agent based systems can augment pure generative language models with knowledge. 
This is crucial in building knowledge-intensive dialog systems.

According to the results in Table \ref{tab:main}, the RAG based systems and the agent based systems perform on par with each other.
As both the RAG based and agent based systems are competitive in building knowledge-intensive dialog systems, it is interesting to compare the two systems and discuss how to improve these systems in future work. 
First, on the one hand, from Table \ref{tab:ret_result}, we can see that the search APIs in the agent system perform better than the RAG system in the retrieval task.
On the other hand, from Table \ref{tab:api_predict}, we can observe that the agent system suffers from low decision making accuracy, whether by prompting or by 
\modelname{}, while the RAG system has no such limitation.
The combined effect is that the agent systems perform close to the RAG system.
The performance difference between the RAG systems and the agent systems on the knowledge search task and the decision making task may vary under different domains. Therefore, it is suggested to explore both agent systems and RAG systems for a certain real-world application in order to achieve the best performance. 

Second, we examine whether the knowledge pieces provided by the RAG retriever or the agent search API is accurate enough for the language model to generate the ideal responses. The results in Table \ref{tab:ret_result} show that the recall@1 is relatively low for the product and FAQ search, especially the product search, indicating that more efforts should be put into increasing the knowledge retrieval accuracy for both RAG and agent systems. 
Moreover, as shown in Table \ref{tab:oracle_result}, we can find out that using the annotated knowledge instead of the retrieved knowledge in testing can greatly improve the RAG performance, which also emphasizes the importance of accurate knowledge retrieval. 

Finally, we examine whether using the annotated knowledge or using the knowledge retrieved by the retriever in the training process in the RAG-based system will yield better performance. The results in Table \ref{tab:noise_result} show that using the retrieved knowledge in the training stage will greatly improve the performance, as the generator needs to discern whether the retrieved knowledge are correct or not. In testing, the oracle knowledge is not provided, and therefore the ability to discern whether the knowledge provided by the retriever is correct is an important skill for a good generator. 
\vspace{-0.5em}
\section{Conclusion}

In this paper, we propose to finetune the LLMs in the 
dialog systems with domain-specific data, together with the domain-specific external knowledge, which is called knowledge augmented finetuning (\modelname{}). The proposed \modelname{} method aims to teach LLMs how to make use of external knowledge. 
To test the efficacy of the \modelname{} method, we build RAG-based and agent-based dialog systems with the \modelname{} method, leveraging the real-life customer service dataset MobileCS2.
In our experiments, systems using the \modelname{} method achieve substantial performance gains over those using the prompting method, particularly in terms of factual accuracy, which shows the efficacy of \modelname{} in building knowledge-intensive dialog systems. With the \modelname{} method, the model gains improved capability of making use of external knowledge in both RAG-based and agent-based dialog systems. 
Furthermore, we conduct ablation studies on the knowledge usage and accuracy in the systems, which shed light on future work on building dialog systems that can provide more accurate responses.

\bibliographystyle{IEEEbib}

\bibliography{refs}

\end{document}